%% file: main-arxiv-v2.tex
\documentclass[10pt,twocolumn,letterpaper]{article}

\usepackage{multirow}
\usepackage{multicol}
\usepackage{booktabs}
\usepackage{amsmath, amssymb}
\usepackage{type1cm}
\usepackage{xcolor,colortbl}
\usepackage{amsmath}
\usepackage{tabularx}

\usepackage{subcaption}
\usepackage{comment}
\usepackage{cuted}
\usepackage{wrapfig}

\usepackage[pagenumbers]{cvpr} %

\input{preamble}

\definecolor{cvprblue}{rgb}{0.21,0.49,0.74}
\usepackage[pagebackref,breaklinks,colorlinks,allcolors=cvprblue]{hyperref}

\definecolor{LightRed}{HTML}{FFE0E0}

\def\paperID{2032} %
\def\confName{CVPR}
\def\confYear{2026}

\title{Learning to Assist: Physics-Grounded Human-Human Control \\
via Multi-Agent Reinforcement Learning}
\author{
Yuto Shibata$^{1,2, 3}$ \quad
Kashu Yamazaki$^{1}$ \quad
Lalit Jayanti$^{1}$ \quad \\
Yoshimitsu Aoki$^{2, 3}$ \quad
Mariko Isogawa$^{2, 3}$ \quad
Katerina Fragkiadaki$^{1}$\\
{ 
      \small
      $^{1}$Carnegie Mellon University
    } 
    {
      \small
      $^{2}$Keio AI Research Center
      \quad\quad
      $^{3}$Keio University
    } \\
    {
        \small
        \url{https://yutoshibata07.github.io/AssistMimic/}
    }
}

  \definecolor{myorange}{RGB}{255,165,0}
\begin{document}
\maketitle

\begin{strip}
  \centering
    \vspace{-15mm}
    \includegraphics[width=1.0\linewidth]{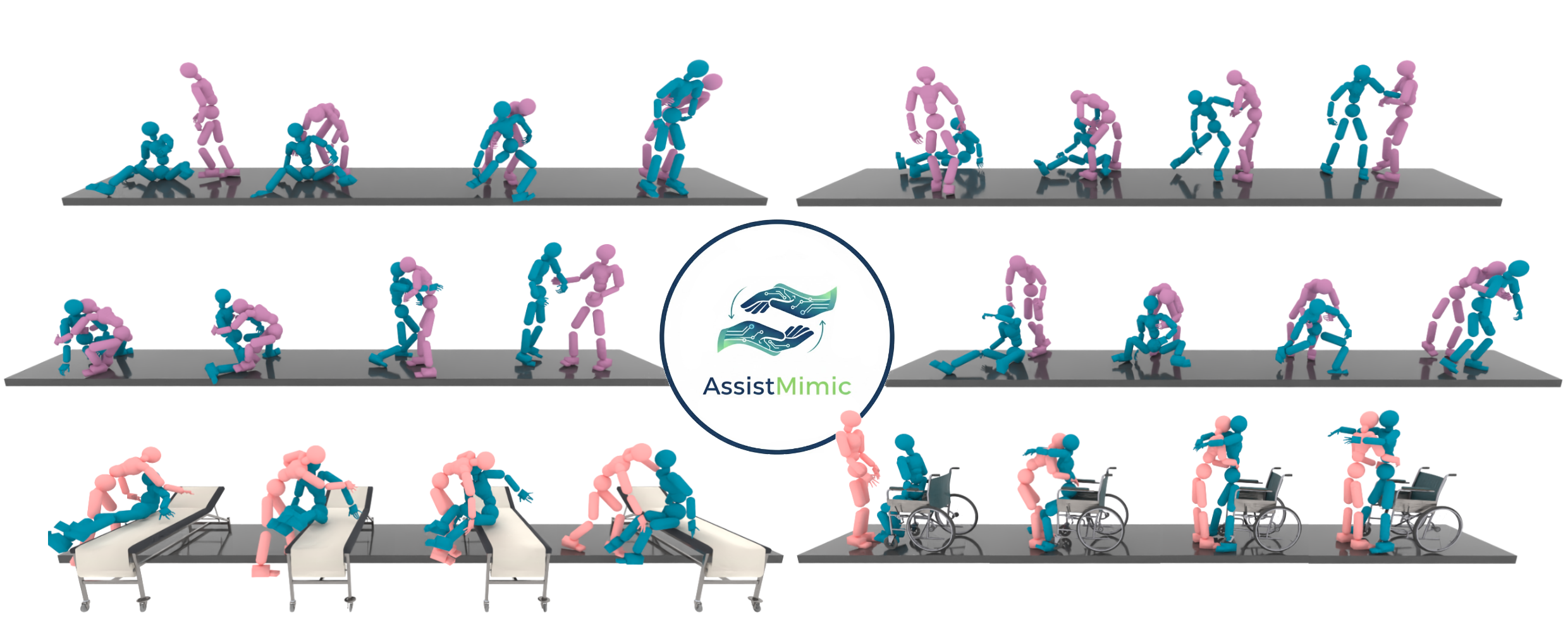}
    \captionof{figure}
    {
    \textbf{\model{}}: We propose a multi-agent RL framework capable of learning robust Supporter and Recipient policies from noisy, close-proximity motion sequences. By leveraging single-person motion priors, a novel recipient-adaptive reference retargeting mechanism, and contact-promoting rewards, \model{} becomes the first physics-based controller to successfully track such complex, high-contact reference motions. Snapshots are arranged chronologically from left to right. 
    }
    \label{fig:teaser}
\end{strip}

\input{sec-cameraready/00_abstract}

\input{sec-cameraready/01_intro}

\input{sec-cameraready/02_related_work}

\input{sec-cameraready/03_method}

\input{sec-cameraready/04_expk}

\input{sec-cameraready/05_conclusion}

\newpage
\paragraph{Acknowledgments.} 
This work was partially supported by JSPS KAKENHI Grant Number 24K22296 and 25H01159, and by JST BOOST, Japan Grant Number JPMJBS2409.

{
    \small
    \bibliographystyle{ieeenat_fullname}
    \bibliography{main}
}

\clearpage
\appendix
\onecolumn
\input{supp-arxiv2}

\end{document}

%% file: preamble.tex
\newcommand{\model}{AssistMimic}

%% file: sec-cameraready/00_abstract.tex
\begin{abstract}
Humanoid robotics has strong potential to transform daily service and caregiving applications. Although recent advances in general motion tracking within physics engines (GMT) have enabled virtual characters and humanoid robots to reproduce a broad range of human motions, these behaviors are primarily limited to contact-less social interactions or isolated movements. Assistive scenarios, by contrast, require continuous awareness of a human partner and rapid adaptation to their evolving posture and dynamics.
In this paper, we formulate the imitation of closely interacting, force-exchanging human–human motion sequences as a multi-agent reinforcement learning problem. We jointly train partner-aware policies for both the supporter (assistant) agent and the recipient agent in a physics simulator to track assistive motion references. To make this problem tractable, we introduce a partner policies initialization scheme that transfers priors from single-human motion-tracking controllers, greatly improving exploration. We further propose dynamic reference retargeting and contact-promoting reward, which adapt the assistant’s reference motion to the recipient’s real-time pose and encourage physically meaningful support.
We show that \model{} is the first method capable of successfully tracking assistive interaction motions on established benchmarks, demonstrating the benefits of a multi-agent RL formulation for physically grounded and socially aware humanoid control. 
\end{abstract}

%% file: sec-cameraready/01_intro.tex
\section{Introduction}
Recent advances in humanoid control and character animation have been driven by motion imitation from human motion capture data \cite{chen2025gmt,xu2025intermimic,zhao2025resmimic,Luo_2023_ICCV,tessler2024maskedmimic}. Replacing hand-engineered task rewards \cite{humanoidparkour} with motion imitation provides a powerful and general reward formulation that minimizes reward design effort while naturally defining task objectives for both robots and virtual characters. Retargeting human motion to humanoid robots and training tracking-based controllers have enabled agile, robust, and disturbance-resistant behaviors \cite{ji2024exbody2,he2025hover,chen2025gmt,zhang2025any2track,xu2025parc}.

Despite these advances, existing tracking-based controllers primarily focus on single-person motion sequences~\cite{Luo_2023_ICCV}. 
In contrast, interactive behaviors involving close contact and force exchange, which are central to assistive and caregiving scenarios, remain a major open challenge.
Learning such behaviors requires not only reproducing reference motions but also continuously adapting to a partner's changing dynamics. 

Prior work typically addresses multi-agent interaction with a \textit{kinematic replay} strategy: a pre-trained single-agent controller generates the recipient's motion in isolation, which is then played back while the assistive agent learns to react \cite{ji2025towards,liu2024physreaction}. 
However, this approach cannot be applied to scenarios in which multiple agents move simultaneously while influencing each other. Indeed, in tightly coupled assistive motions, the recipient’s behavior cannot be computed independently, and decoupling the learning of partner control policies breaks physical consistency.

We address this challenge with \model{}, a multi-agent reinforcement learning (MARL) framework for learning physics-aware, tracking-based controllers for close human–human interactions. Unlike prior approaches that rely on kinematic replay, \model{} jointly trains policies for both partners within a MARL formulation~\cite{yu2022surprising,gao2024coohoi}. This joint optimization removes the need to ``freeze" one agent's motion, allowing the assistant and recipient to co-adapt during training and enabling rich bidirectional coordination, particularly when the recipient becomes physically unstable. 

\begin{figure}[t]
\centering
\centerline{\includegraphics[width=1.0\linewidth]{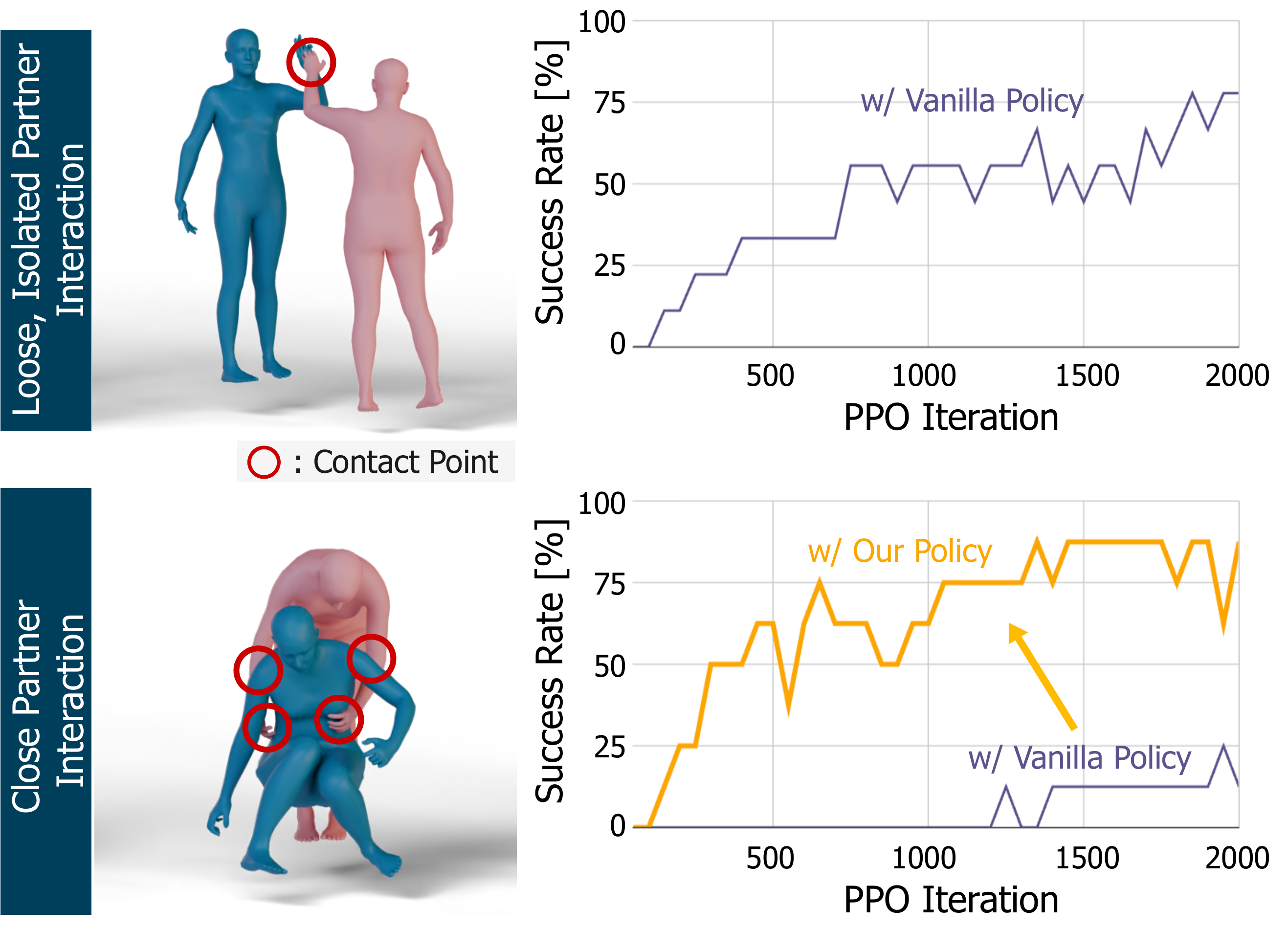}}
\vspace{-2mm}
\caption{
Learning contact-rich assistive behaviors is substantially more difficult in the close-contact interactions that we target (bottom) than in contact-less social interactions (top) or isolated motions (\textcolor{gray}{gray} SR curve). \model{} addresses these challenges, achieving successful imitation for the first time, as shown in the improved \textcolor{orange}{orange} SR curve. 
}
\label{fig:overview}
\vspace{-5mm}
\end{figure}

Yet, reinforcement learning for reactive control in physically entangled interactions is highly challenging.
As shown in Figure~\ref{fig:overview}, unlike the contact-less social interactions or isolated motions examined in prior work such as high-fives (top), contact-rich assistive behaviors (bottom) require precise spatial alignment and tightly timed force application; even minor errors in contact location or force magnitude can destabilize the recipient. The difficulty is compounded by severe occlusion in close-range human–human motion capture, yielding noisy and inaccurate reference trajectories. These factors make the MARL training unstable. 

\model{} facilitates exploration in MARL through three core components. It initializes partner policies from single-person tracking controllers to leverage strong locomotion priors, uses dynamic reference retargeting to maintain stable close-contact interactions, and incorporates contact-promoting rewards to enable learning from noisy or incomplete demonstrations. Together, these innovations enable stable and physically realistic imitation of interactive assistive motions in simulation, leading to successful learning, as shown by the orange curve in Figure~\ref{fig:overview}.

We evaluate \model{} on the Inter-X \cite{xu2024inter} and HHI-Assist \cite{saadatnejad2025hhi} datasets, both of which feature tightly coupled, support-oriented human–human interactions. Across these benchmarks, \model{} is the only method to track closely interacting, force exchanging human motions, achieving substantially higher task success rates (Inter-X: 83\%, HHI-Assist: 73\%) and demonstrating strong robustness to unseen disturbances such as stumbles and load shifts. Empirically, \model{} accurately reproduces reference motions across diverse assistive scenarios—including recipients lying on the floor, seated in chairs, or resting in beds—where the assisting agent learns to provide appropriate physical support and lifting assistance. To objectively assess assistance quality, we reduce the recipient’s PD gains and maximum torque limits to constrain self-stabilization, thereby isolating the contribution of the learned assistive controller.

In summary, our key contributions are as follows:
\begin{itemize}
    \item \textbf{Formulation:} We cast physics-based human--human motion imitation as a multi-agent reinforcement learning (MARL) framework for learning physics-enabled tracking controllers, enabling motion imitation for caregiving and assistive tasks that require reactive force exchange.
    \item \textbf{Methodology:} We further introduce motion prior initialization, dynamic reference retargeting, and contact-promoting reward schemes that make multi-agent RL tractable and effective in high-contact settings.
    \item \textbf{Evaluation:} We conduct extensive experiments and ablations that quantify each component's contribution to learning efficiency, imitation fidelity, and assistive stability.
\end{itemize}

%% file: sec-cameraready/02_related_work.tex
\section{Related Work}

\paragraph{Physics-Based Human Motion Synthesis.}
Early work on physics-based character control, such as DeepMimic~\cite{peng2018deepmimic}, demonstrated that reinforcement learning (RL) can reproduce diverse motion sequences within physically realistic simulators~\cite{park2019,xue2022ase}. Follow-up frameworks, including AMP~\cite{peng2021amp} and PHC~\cite{Luo_2023_ICCV}, expanded this paradigm to generate more stable and diverse behaviors through adversarial rewards and goal-conditioned training~\cite{merel2019NeuralPM,yuan2021simpoe}. More recent approaches, such as PhysDiff~\cite{yuan2023physdiff}, integrate diffusion-based generative models with physics simulation to improve realism and stability. However, these efforts primarily focus on single-agent motion and remain limited in modeling the reactive forces and bidirectional dependencies inherent in multi-agent interactions targeted in our work.

\paragraph{Human–Human and Human–Robot Interaction Synthesis.}
The study of human–human interaction synthesis has evolved rapidly, with recent diffusion-based~\cite{liang2024intergen,xu2024regennet,shafir2024human} and transformer-based~\cite{xu2023actformer,Chopin2023} architectures enabling context-aware multi-agent motion generation. While these models effectively capture social cues and motion continuity, they are typically trained in kinematic space and thus lack physical grounding. 
The recent Human-X framework~\cite{ji2025towards} takes a step toward bridging this gap by integrating an autoregressive diffusion planner with a physics-tracking policy to achieve real-time, physically plausible interactions across human–avatar, human–humanoid, and human–robot domains. Yet, its motion control remains largely reactive and open-loop, relying on pretrained diffusion priors rather than continuous bidirectional adaptation, in which both agents generate motions conditioned on each other's states.

\paragraph{Multi-Agent Reinforcement Learning for Physical Interaction.}
Multi-agent reinforcement learning (MARL) has shown promise for modeling interactive behaviors in physically coupled settings~\cite{yu2022surprising,gao2024coohoi}. Within humanoid control, however, most MARL methods focus on competitive or cooperative games rather than embodied physical contact~\cite{bansal2018marl}. In contrast, our work formulates close-contact human–human motion imitation as a fully coupled MARL problem, where both agents are trained jointly to produce physically consistent, reactive coordination. This design extends prior diffusion-based frameworks~\cite{ji2025towards,liu2024physreaction} by incorporating force feedback and partner-conditioned rewards, enabling adaptive support and assistive behaviors.

%% file: sec-cameraready/03_method.tex
\begin{figure*}[t]
\centering
\centerline{\includegraphics[width=0.95\linewidth]{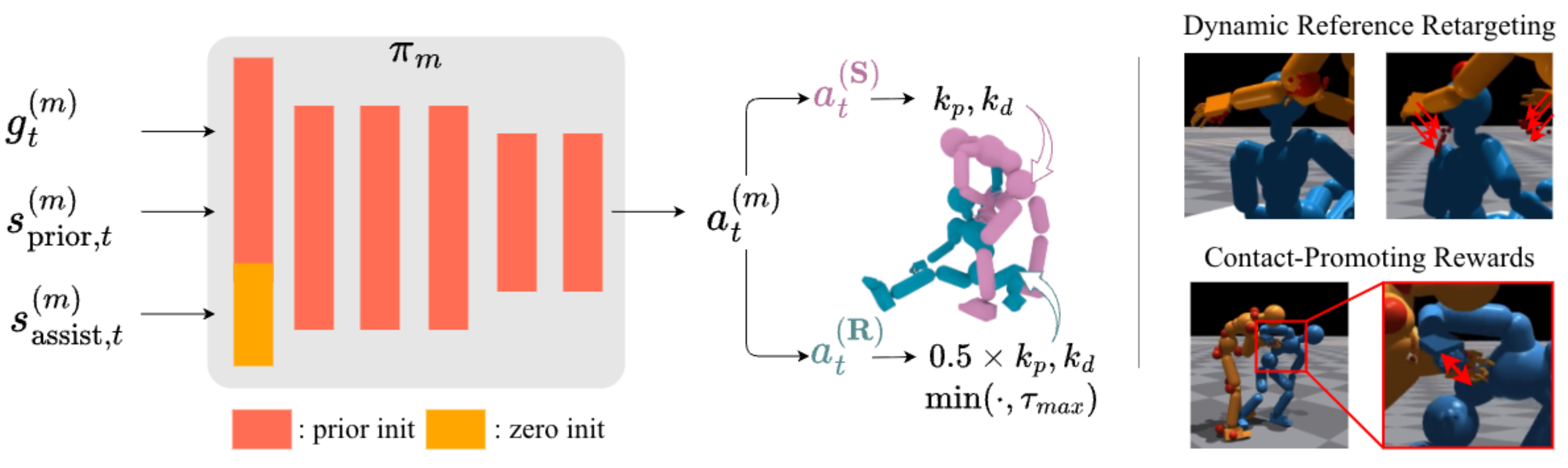}}
\vspace{-4mm}
\caption{\textbf{Overview of \model{}}. We train tracking-based humanoid control policies for both the recipient
and the supporter, optimizing them to imitate a paired reference motion sequence. Our architecture builds on the single-agent tracking framework of PHC~\cite{Luo_2023_ICCV}, extending it with partner-aware state inputs and augmenting standard imitation rewards with recipient-aware reference retargeting and contact-incentivizing reward terms. 
The policy $\pi_m$ takes as input the proprioceptive state $s_{\text{prior},t}^{(m)}$, the assistive state $s_{\text{assist},t}^{(m)}$ (the interaction context and partner information), and the goal $g_t^{(m)}$ to output the action $a_t^{(m)}$.
}
\label{fig:pipeline}
\vspace{-4mm}
\end{figure*}

\section{Method}

The overall architecture of \model{} is illustrated in Figure ~\ref{fig:pipeline}. It formulates human-human interactive motion imitation as a Multi-Agent Reinforcement Learning (MARL) problem. Our framework focuses on assistive scenarios involving two distinct agents: an \textit{assistant} who provides physical support, and a \textit{recipient} who requires assistance. To simulate physical impairment, we impose specific constraints on the recipient's dynamic parameters, including reduced maximum joint torques and lowered Proportional-Derivative (PD) gains ($k_p, k_d$). 
Both agents learn tracking policies that are optimized for reference motion fidelity under these physics constraints. Each policy perceives the environment through a partner-aware, ego-centric state space. To bootstrap learning, we initialize both agents using a pre-trained motion prior, allowing them to inherit common coordination skills before being trained to acquire role-specific behaviors.

\begin{table}[t]
\centering
\caption{Physical limitations applied to the recipient.}
\vspace{-2mm}
\label{tab:phisical_limitations}
\setlength{\tabcolsep}{4pt}
\renewcommand{\arraystretch}{1.05}
\small
\begin{tabularx}{\columnwidth}{l X c c}
\toprule
\multirow{2}{*}{\textbf{Dataset}} 
& \multirow{2}{*}{\textbf{Joint group}} & \textbf{Scale} & \textbf{Max $\tau$}\\
& & $k_p$/$k_d$ & [Nm] \\
\midrule
\multirow{2}{*}{Inter-X~\cite{xu2024inter}} & Lower body & \multirow{2}{*}{0.5} & \multirow{2}{*}{80} \\
& (hips, knees, ankles, toes) & & \\
\midrule
\multirow{5}{*}{HHI-Assist~\cite{saadatnejad2025hhi}} 
& Lower body & \multirow{2}{*}{0.5} & \multirow{2}{*}{80} \\
& (knees, ankles, toes) & & \\
& Upper body & \multirow{2}{*}{0.5} & \multirow{2}{*}{40} \\
& (torso, chest, spine) & & \\
& Hips & 0.5 & 20 \\
\bottomrule
\end{tabularx}
\vspace{-6mm}
\end{table}

\subsection{Problem Formulation}
\label{sec:problem_setup}

\paragraph{Multi-Agent MDP with Asymmetric Dynamics.}
We model close-range assistance as a finite-horizon Markov Decision Process (MDP) defined by the tuple $\mathcal{M}=(\mathcal{S},\mathcal{A},\mathcal{P}_{\kappa},r,\gamma, T)$, with horizon $T$ and discount factor $\gamma \in (0,1)$. Let the two humanoid agents be indexed by $m \in \{\mathrm{S}, \mathrm{R}\}$ corresponding to the \underline{S}upporter agent and \underline{R}ecipient agent. The state space $\mathcal{S}$ and action space $\mathcal{A}$ represent the joint configurations and actuations of both agents.
At each timestep $t$, an action $\mathbf{a}_t=(\mathbf{a}_t^{(\mathrm{S})},\mathbf{a}_t^{(\mathrm{R})}) \in \mathcal{A}$ is sampled independently from the  policies:
\begin{equation}
\mathbf{a}_t^{(m)} \sim \pi_{m}(\cdot \mid \mathbf{s}_t^{(m)}; \boldsymbol{\phi}_{m}), \quad m\in\{\mathrm{S},\mathrm{R}\},
\end{equation}
where $\boldsymbol{\phi}_{m}$ denotes the policy parameters and $\mathbf{s}_t^{(m)}$ denotes the agent-centric observation.

The environment transition dynamics $\mathcal{P}_{\kappa}: \mathcal{S} \times \mathcal{A} \to \mathcal{S}$ encompass the physical interactions between the agents. To model physical impairment, the recipient's dynamics are parameterized by $\kappa = (k_p, k_d, \tau_{max})$. We explicitly constrain the recipient's Proportional-Derivative (PD) controller gains ($k_p, k_d$) and maximum joint torques $\tau_{max}$ (details in Table~\ref{tab:phisical_limitations}). These reduced parameters weaken intrinsic stabilization, thereby destabilizing the recipient's motion and necessitating external support to prevent falls.

\paragraph{Reference Motion and Objective.}
For each agent $m$, we define a reference motion sequence $\{\hat{\mathbf{q}}^{(m)}_t\}_{t=1}^T$. The reference state $\hat{\mathbf{q}}^{(m)}_t$ is a tuple comprising joint rotations $\hat{\boldsymbol{\theta}}_t$, joint positions $\hat{\mathbf{p}}_t$, joint angular velocities $\hat{\boldsymbol{\omega}}_t$, and joint linear velocities $\hat{\mathbf{v}}_t$:
\begin{equation}
\hat{\mathbf{q}}^{(m)}_t \triangleq (\hat{\boldsymbol{\theta}}^{(m)}_t, \hat{\mathbf{p}}^{(m)}_t, \hat{\boldsymbol{\omega}}^{(m)}_t, \hat{\mathbf{v}}^{(m)}_t).
\end{equation}
We denote reference quantities with hats $(\hat{\cdot})$ and simulated states without. 
The goal is to learn optimal policy weights $(\boldsymbol{\phi}_{\mathrm{S}}^*, \boldsymbol{\phi}_{\mathrm{R}}^*)$ that maximize the expected discounted return:
\begin{equation}
J(\pi_{\mathrm{S}},\pi_{\mathrm{R}}) = \mathbb{E}_{\tau \sim \mathcal{P}{\kappa}}\left[\sum_{t=0}^{T-1} \gamma^{t} (r_t^{(\mathrm{S})} + r_t^{(\mathrm{R})}) \right].
\end{equation}
The base per-agent reward encourages fidelity to the reference motion. We define this tracking objective using a weighted distance function $D(\cdot, \cdot)$ that accounts for the differing units of state components:
\begin{equation}
\label{eq:standard_reward}
r_{\mathrm{track}}^{(m)} = \exp\left(-D\left(\hat{\mathbf{q}}^{(m)}_t, \mathbf{q}^{(m)}_t\right)\right).
\end{equation}
For the recipient ($m=\mathrm{R}$), the task reward consists of tracking,
power, and assist stability terms; the full decomposition is provided in Appendix~E.
However, for the supporter ($m=\mathrm{S}$), we modify this objective to satisfy interaction requirements. First, as detailed in Section~\ref{sec:assist-hand-align}, the supporter's hand reference trajectories are dynamically retargeted relative to the recipient's body to maintain valid relative positioning. Second, as described in Section~\ref{sec:contact_reward}, the standard hand-tracking reward terms are replaced by contact-incentivizing objectives when in close proximity to the recipient, prioritizing functional support over strict kinematic adherence.

\subsection{Goal-conditioned tracking policy architecture} 
We implement the supporter ($\pi_S$) and recipient ($\pi_R$) policies using a symmetric, goal-conditioned policy architecture. 
\textbf{\model{} builds on the insight that single-human motion priors can  greatly ease exploration in RL for imitating human-human interactions.  }
These priors can supply essential locomotion skills—like standing and walking—so the policy need not relearn basic dynamics from scratch.  We thus adapt the PHC \cite{Luo_2023_ICCV} tracking-based single-person policy as our prior policy $\pi_\mathrm{prior}$ and augment its architecture by extending the input layer to incorporate interaction-aware features.
The input space of the original PHC is expanded to include the assistive state, denoted as $s_\mathrm{assist}$. Consequently, the policy input is defined as the tuple $(s_\mathrm{prior}, s_\mathrm{assist}, g)$. In this symmetric setup, $s_{prior}$ always represents the ego agent's state (supporter for $\pi_S$, recipient for $\pi_R$), while $s_\mathrm{assist}$ represents the interaction context and partner information relative to the ego agent. 
Specifically, $s_\mathrm{prior}$ is proprioception that includes joint rotations $\boldsymbol{\theta}_t\!\in\!\mathbb{R}^{J\times 6}$, joint positions $\mathbf{p}_t\!\in\!\mathbb{R}^{J\times 3}$, joint angular velocities $\boldsymbol{\omega}_t\!\in\!\mathbb{R}^{J\times 3}$, joint linear velocities $\mathbf{v}_t\!\in\!\mathbb{R}^{J\times 3}$, and root height $h_t\!\in\!\mathbb{R}^{1}$. 

\begin{equation}
    s_{\mathrm{prior}, t}^{(m)}=\{\boldsymbol\theta_t^{(m)},\mathbf{p}_t^{(m)},\boldsymbol\omega_t^{(m)},\mathbf{v}_t^{(m)}\, h_t^{(m)}\}
\end{equation}
The goal $g_t^{(m)}$ is defined as a set of reference delta state calculated between the current state and the reference motion state in the next timestep.

The assistive state $s_\mathrm{assist}$ encodes partner-aware information. It consists of the partner observation $\mathbf{o}^\mathrm{(\bar{m})}_t\in\mathbb{R}^{6+J\times(3+3+6)+2\times J\times 3}$ (containing the partner's gravity-aligned root rotation, joint positions/velocities/rotations, and joint positions relative to the ego-agent's wrists), the partner's hand contact state $\boldsymbol c^\mathrm{(\bar{m})}_t \in \{0,1\}^{12}$, the ego-agent's contact state $\boldsymbol c^{(m)}_t \in \{0,1\}^{12}$, self-force $\;\mathbf{f}^{(m)}_t \in \mathbb{R}^{14 \times 3}$ (3d contact forces on elbows, wrists, and fingertips) and the own previous action $\mathbf{a}^{(m)}_{t-1} \in\mathbb{R}^{J \times 3}$.

\begin{equation}
    s_{\mathrm{assist}, t}^{(m)}=\{\mathbf{o}^\mathrm{(\bar{m})}_t\, \boldsymbol c^\mathrm{(\bar{m})}_t, \boldsymbol c^\mathrm{(m)}_t, \mathbf{f}^\mathrm{(m)}_t, \mathbf{a}^\mathrm{(m)}_{t-1}\}
\end{equation}
Here, $\mathbf{c}_t$ is a binary indicator that activates when the distance between the agent's hand and the partner's hand falls below a threshold, and the magnitude of the agent's hand force exceeds a set limit. Both self-forces $\mathbf{f}^{(m)}_t$ and partner observations $\mathbf{o}^\mathrm{(\bar{m})}_t$ are computed in the agent-centric ego frame.

\vspace{-3mm}

\paragraph{Weight Initialization from Single Person Motion Prior.}
We initialize both the supporter ($\pi_S$) and recipient ($\pi_R$) policies using parameters from the pre-trained prior $\pi_\mathrm{prior}$ of single-person tracking base policy. However, as the \model{} requires the additional assistive state $s_\mathrm{assist}$, the input layer dimension increases. To account for this while preserving the prior's initial behavior, we employ a zero-padding initialization strategy. The weights corresponding to the original proprioceptive state $s_\mathrm{prior}$ are copied from the prior, while the weights corresponding to the new assistive features are initialized to zero. Formally, let $\mathbf{W}^\mathrm{input}_\mathrm{prior} \in \mathbb{R}^{H \times D_{p}}$ be the input weights of the prior network. The new input weight matrix $\mathbf{W}^\mathrm{input}_\mathrm{new} \in \mathbb{R}^{H \times (D_{p} + D_{a})}$ is constructed as:

\vspace{-3mm}

\begin{equation}
    \mathbf{W}^\mathrm{input}_\mathrm{new} = \left[\mathbf{W}^\mathrm{input}_\mathrm{prior} \mid \mathbf{0}\right],
\end{equation} 
where $\mathbf{0} \in \mathbb{R}^{H \times D_{a}}$ is a zero matrix matching the dimension of the assistive state inputs.

\subsection{Dynamic Reference Retargeting}
\label{sec:assist-hand-align}
In close-contact scenarios, effective assistance must adapt to the recipient’s state. Simply tracking a fixed reference trajectory fails whenever the simulated recipient deviates from the motion capture sequence, as the spatial configuration required for support quickly breaks down. To address this, we introduce a dynamic reference retargeting mechanism that continuously adjusts the assistant’s hand targets to preserve their intended relative position with respect to the recipient’s body.

This adaptation is triggered based on the agents’ spatial proximity.  Let $\mathcal{G}_t$ denote the gating indicator:
\begin{equation}
\mathcal{G}_t = \mathbb{I}\left(\left|\left| \mathbf{p}^{(S)}_{root, t} - \mathbf{p}^{(R)}_{root, t} \right|\right|_2 \le \tau_{\mathrm{dist}}\right)
\end{equation}
which activates retargeting when the assistant’s and recipient’s root positions are sufficiently close.

For each wrist $i \in \{L, R\}$ for \underline{L}eft and \underline{R}ight, let $\mathcal{H}_i$ denote the set of associated hand joints (wrist and fingers). We first identify the \textit{anchor} joint on the recipient's body—the specific anatomical site the supporter intends to assist—by finding the nearest recipient joint in the canonical reference space (denoted by $\hat{\cdot}$):
\begin{equation}
k_{i,t}^{*} = \arg\min_{k \in \mathcal{J}_R} \left|\left| \hat{\mathbf{p}}^{(R)}_{k, t} - \hat{\mathbf{p}}^{(S)}_{i, t} \right|\right|_2. 
\end{equation}
We compute the desired relative offset $\Delta\hat{\mathbf{p}}_{h_i,t} = \hat{\mathbf{p}}^{(S)}_{h_i,t} - \hat{\mathbf{p}}^{(R)}_{k^*, t} \;\;\forall\, h_i\in\mathcal{H}_i$. and transfer it to the simulated recipient state $\mathbf{p}^{(R)}_{k^*,t}$. The updated tracking target becomes: $\hat{\mathbf{p}}^{(S)}_{h_i, t} = \mathbf{p}^{(R)}_{k^*, t} + \Delta\hat{\mathbf{p}}_{h_i,t}$ iff $\mathcal{G}_t = 1$. This procedure ensures that the assistant’s hands track a physically meaningful interaction point relative to the recipient’s current pose, rather than a fixed location in global space.

\subsection{Contact-Promoting  Rewards}
\label{sec:contact_reward}
Effective supportive motion requires precise spatiotemporal placement of the hand and appropriate force exertion. However, raw motion capture data frequently suffer from occlusion noise, making hand trajectories unreliable. Consequently, strict reference tracking can hinder effective support or lead to inadvertent collisions. To mitigate this, we introduce contact-incentivizing rewards when  the recipient's body is in close proximity to the assistant's hands. 

Specifically, we index the supporter's wrists by $i \in \{L, R\}$ and the recipient's upper-body joints by $j \in \mathcal{U}$. 
We define the minimum wrist-to-recipient distance $d_{i, t}$ and a binary proximity indicator $\chi_{i, t}$:

\begin{equation}
d_{i, t} = \min_{j\in\mathcal{U}} \left|\left| \mathbf{p}^{(S)}_{i, t} - \mathbf{p}^{(R)}_{j,t} \right|\right|_2, \; \chi_{i, t} = \mathbb{I}(d_{i, t} \le d_{\mathrm{th}}).
\end{equation}

When the supporter’s hands are distant from the recipient ($\chi_{i, t}=0$), we utilize the standard tracking reward (Eq. \ref{eq:standard_reward}). When in close proximity ($\chi_{i, t}=1$), we suppress the tracking penalty, assuming zero tracking error, and activate the contact-promotion term:

\begin{equation}
\hspace{-5mm}
\begin{aligned}
r_{\mathrm{track}_i}^{(S)} 
\hspace{-1mm}
=
\hspace{-1mm}
\left\{
\begin{aligned}
 & \exp\left(-D\left(\hat{\mathbf{q}}^{(S)}_{i, t}, \mathbf{q}^{(S)}_{i, t}\right)\right)
   && \hspace{-1mm} \text{if }\chi_{i, t} = 0,\\[2pt]
 & \beta f_{i, t} \exp(-\alpha d_{i, t}) + b_{\text{contact}}
   && \hspace{-1mm} \text{if }\chi_{i, t} = 1,
\end{aligned}
\right.
\end{aligned}
\end{equation}

where $\alpha, \beta$ are hyperparameters for scaling, 
$b_{\mathrm{contact}}$ is a sparse bonus awarded upon establishing contact, and $f_{i, t}$ is the measured finger-contact forces aggregated with a safety-aware soft saturation function:

\begin{equation}
f_{i, t} = \sum_{\ell\in\mathcal{H}_{i} \setminus i}
\min\left(\exp(||\mathbf{f}_{\ell,t}||_2 - f_{\mathrm{th}}), 1\right)
\end{equation}

where $\mathbf{f}_{\ell, t}$ is the contact force at finger joint $\ell$ and $f_{\mathrm{th}}$ is the force threshold.

\paragraph{Implementation details}

We cluster motions based on subject ids, and train a single ``specialist" policy per cluster with PPO~\cite{schulman2017proximal}. We distill specialist policies into a ``generalist" policy to be able to handle more diverse reference motions,
using DAgger~\cite{ross2011reduction}. 
For both specialist and generalist training, we apply tight early termination~\cite{tessler2025maskedmanipulator} with a 0.25\,m pose-deviation threshold. For initialization, we adopt \textit{Physical State Initialization} (PSI~\cite{xu2025intermimic}): instead of sampling from noisy reference trajectories that often contain penetrations, we seed episodes from states collected in recent rollouts. Other hyperparameters are detailed in Appendix~D.

%% file: sec-cameraready/04_expk.tex
\section{Experiments}

Our experiments aim to answer the following questions: \textbf{(1)} How well can \model{} specialist and generalist policies track closely interacting humans in force-exchanging assistive motions? \textbf{(2)} How important is the formulation of physics-based human-human imitation as a multi-agent RL problem, over single-agent RL conditioned on a trajectory replay of the recipient's trajectory? 
\textbf{(3)} What is the contribution of various components of our framework?

\noindent
\textbf{Datasets}
We evaluate \model{} in its ability to track close-range, contact-rich assistive motion reference trajectories included in the following two datasets:
\begin{itemize}
\item  \textbf{Inter-X}~\cite{xu2024inter} 
This is the largest two-person HHI dataset and stores motions in the SMPL-X~\cite{SMPL-X:2019} format. Unlike prior work that focuses only on contact-less social interactions, such as high-five~\cite{liu2024physreaction,ji2025towards}, we extracted 30 paired “Help-up” motions to cover a wide range of assistive strategies. 
These motions are diverse in both approach and contact patterns, spanning directions such as front and back, and support types from single-arm grasping to two-hand shoulder support.
\item \textbf{HHI-Assist}~\cite{saadatnejad2025hhi} This dataset captures caregiving motions on a bed or chair. It includes specialized actions such as supporting the recipient’s shoulder while lifting them from the bed, or stabilizing the knees and rotating the body to a seated posture. We use 10 representative caregiving motion clips (details in Appendix~D). 
\end{itemize}

\noindent
\textbf{Evaluation settings}
We evaluate \model{} and its baselines under four scenarios:
(i) \textbf{Specialist policy evaluation.} Both our method and the baselines are trained on approximately 8--10 motion clips, grouped by subject, to assess performance in a specialist setting. (ii) \textbf{Generalist policy evaluation.}  As Inter-X includes a broad range of get-up and support strategies, we train and evaluate a single policy on 30 diverse clips to test whether \model{} can reproduce a wide variety of support behaviors with one unified controller.
(iii) \textbf{Support under unseen dynamics.}  We assess robustness to recipient dynamics not seen during training by modifying physical parameters at test time: halving the PD gains from the training phase, increasing body mass to $1.2\times$ or $1.5\times$, and reducing maximum hip torque to $0.5\times$ the training value.
(iv) \textbf{Tracking generated kinematic trajectories.}  We apply our tracking policy to outputs from a generative model that synthesizes two-person interactions.

\noindent
\textbf{Baselines} \
To the best of our knowledge, \model{} is the first method to learn tracking controllers for closely interacting, force-exchanging human--human motion sequences. We evaluate our method against the following paradigms to justify our multi-agent formulation:

\begin{itemize}
\item \textbf{Phys-Reaction}~\cite{liu2024physreaction}: This method employs an isolated rollout strategy using a single-agent controller~\cite{Luo_2023_ICCV}. However, in assistive scenarios, the recipient's motion is often physically unrealizable without external support. In our experiments, it failed to produce any stable recipient trajectories during the initial tracking phase, making it inapplicable as a direct baseline.

\item \textbf{Kinematic-Recipient}: Following the approach of Human-X~\cite{ji2025towards}, which trains a reactor against a kinematically replayed actor, we implement a baseline where the recipient's motion is a fixed kinematic replay. 

\item \textbf{Frozen-Recipient (Decoupled Learning)}: We first fine-tune a pre-trained single-agent tracker~\cite{Luo_2023_ICCV} on the recipient's motions and then freeze its parameters. We then train the assistant to coordinate with this pre-trained but non-adaptive recipient. This assesses whether sequential, decoupled training is sufficient for mutual coordination.

\item \textbf{Ablated Variants}: We remove key components of \model{}---specifically weight initialization, dynamic reference retargeting, and contact-promoting rewards---to assess their individual contributions.
\end{itemize}

\begin{table*}[t]
\centering
\caption{Evaluation of specialist policies on the \textbf{Inter-X} dataset. }
\vspace{-3mm}
\label{tab:interx_expert}
\setlength{\tabcolsep}{6pt}
\renewcommand{\arraystretch}{1.1}
\small
\begin{tabular}{l c c c c}
\toprule
\multirow{2}{*}{Method} 
& \multicolumn{2}{c}{seen dynamics} 
& \multicolumn{2}{c}{unseen dynamics} \\
\cmidrule(lr){2-3} \cmidrule(lr){4-5}
& SR($\uparrow$)[\%] & MPJPE($\downarrow$)[mm]
& Mass $\times 1.2$ (SR ($\uparrow$)) 
& $K_p/K_d \times 0.5$ (SR ($\uparrow$)) \\
\specialrule{.12em}{.25em}{.25em}

Sequential Training     & 62.4    &   92.3    & 49.9 & 50.5 \\
\rowcolor{pink!30}
\model{} 
& 74.9 & 113 & 57.9 & 72.8
\\
($-$) Dynamic Reference Retargeting & 83.4 & 107  & \textbf{73.1} & \textbf{83.3}\\
\multicolumn{1}{l}{\hspace{1.4em}($-$) Contact Promoting Reward} 
& 81.6 & 80.4 
& 66.3
& 77.1
\\
\multicolumn{1}{l}{\hspace{2.2em}($-$) Weight Initialization} 
& 0.0 & 248 & 0.0 & 0.0 \\
\bottomrule
\end{tabular}
\end{table*}

\begin{table*}[t]
\centering
\vspace{-3mm}
\caption{ Evaluation of specialist policies on the \textbf{HHI-Assist} dataset.}
\vspace{-3mm}
\label{tab:expert-comparison}
\setlength{\tabcolsep}{6pt}
\renewcommand{\arraystretch}{1.1}
\small
\begin{tabular}{l c c c c}
\toprule
\multirow{2}{*}{Method} & \multicolumn{2}{c}{seen dynamics } & \multicolumn{2}{c}{unseen dynamics} \\
\cmidrule(lr){2-3} \cmidrule(lr){4-5}
& SR($\uparrow$)[\%] & MPJPE($\downarrow$)[mm]
& Mass $\times 1.5$ (SR ($\uparrow$)) & Max hip torque $\times 0.5$ (SR ($\uparrow$)) \\
\specialrule{.12em}{.25em}{.25em}
\rowcolor{pink!30}
\model{} & 85.8 & 127.0 & \textbf{67.8} 
& \textbf{73.2}
\\
($-$) Dynamic Reference Retargeting & 85.4 & 125 & 49.1 
& 62.9
\\
\multicolumn{1}{@{\hspace{1.4em}}l}{($-$) Contact Promoting Reward} & 97.7 & 89.5 & 56.4 
& 27.7 
\\
\multicolumn{1}{@{\hspace{2.2em}}l}{($-$) Weight Initialization} & 19.1$^{\dag}$ & 364$^{\dag}$ & - & - \\
\bottomrule
\end{tabular}
\vspace{-5mm}
\end{table*}

\noindent
\textbf{Evaluation Metrics}
We evaluate the controller’s tracking quality using the following metrics:
\begin{itemize}
\item Success Rate (SR), which measures the percentage of episodes that end in success.  An episode is marked as successful if, for both the supporter and the recipient, the distance between the simulated body and its reference motion does not exceed a predefined threshold of 0.5 m. 
\item Mean Per Joint Position Error (MPJPE), which measures the average positional deviation between the simulated and reference joint positions for both agents. 
\item COM Stability on HHI-Assist, which requires steady bed support. 
We measure the temporal stability of the recipient's center of mass (COM).
See Appendix~C for details.
\end{itemize}

\begin{table}[t]
\centering
\small
\caption{Evaluation of generalist policy on the \textbf{Inter-X} dataset.}
\vspace{-3mm}
\label{tab:generalist-comparison}
\begin{tabular}{l *{2}{c}}
\toprule
\multirow{1}{*}{Method} 
& Success Rate (\(\uparrow\))[\%] & MPJPE (\(\downarrow\))[mm]\\
\midrule
\model{} & 77.3 & \textbf{132} \\
\rowcolor{pink!30}
($+$) DAgger & \textbf{94.7} & 168 \\
\bottomrule
\end{tabular}
\vspace{-6mm}
\end{table}

\subsection{Specialist Policy Evaluation}
\noindent
\begin{wrapfigure}[8]{r}{0.55\linewidth}
\centering
\vspace{-4mm}
\includegraphics[width=\linewidth]{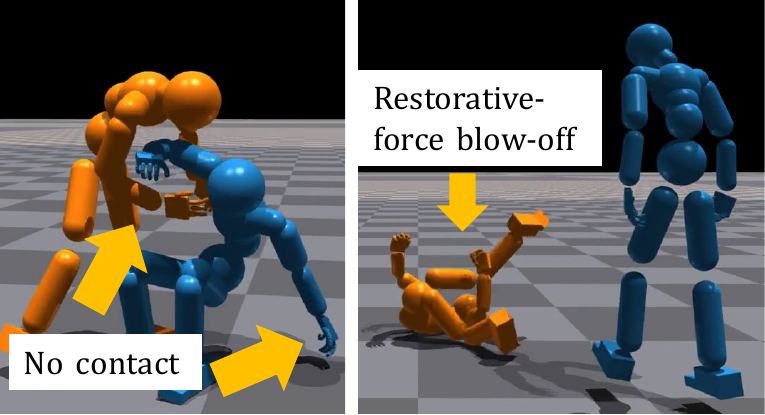}
\vspace{-7mm}
\caption{Failure of Kinematic Baselines.}
\label{fig:recipient_kinematic}
\vspace{-4mm}
\end{wrapfigure}
\noindent
We first show that the recipient kinematic replay is ill-posed for assistive tasks.  
In Figure~\ref{fig:recipient_kinematic} (left), applying the kinematic replay used in Human-X~\cite{ji2025towards} to our scenario causes the recipient to ``stand up'' independently along a pre-defined trajectory, regardless of the presence of support (yellow arrows).
Consequently, the assistant’s actions do not meaningfully affect the partner’s state, preventing the emergence of functional support behaviors. In Figure~\ref{fig:recipient_kinematic} (right), the kinematically fixed poses also cause severe interpenetration, which induces large restorative forces in the physics engine and eventually pushes the characters apart. These failures show that kinematic replay is unsuitable for the recipient in our task.
  
\begin{figure}[t]
\centering
\centerline{\includegraphics[width=1.0\linewidth]{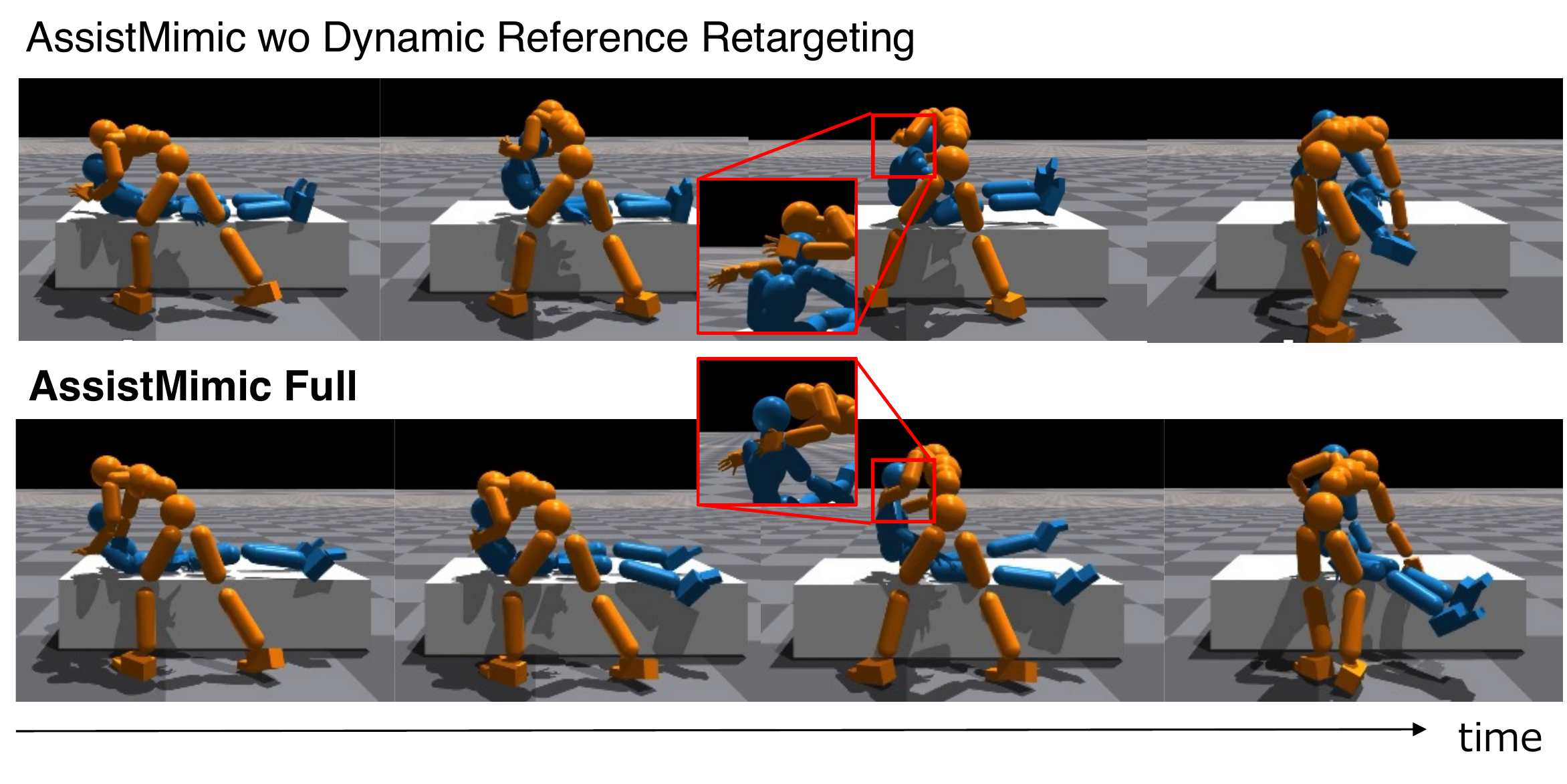}}
\vspace{-5mm}
\caption{Qualitative results on HHI-Assist. Red boxes indicate correct hand adjustment and support.}
\label{fig:hhi-assist-qualitative}
\vspace{-4mm}
\end{figure}

We next demonstrate how our joint MARL formulation addresses these limitations. Tables~\ref{tab:interx_expert} and~\ref{tab:expert-comparison} compare the specialist policies of \model{} against its ablative variants. Table~\ref{tab:interx_expert} also includes the Frozen-Recipient (decoupled sequential learning) approach. Qualitative results on the HHI-Assist dataset are shown in Figure~\ref{fig:hhi-assist-qualitative}. Our analysis yields the following key insights:

\begin{itemize}
\item \textbf{Joint Training vs. Sequential Learning}: As shown in Table~\ref{tab:interx_expert}, the \textit{Sequential Training} baseline is substantially outperformed by our joint MARL formulation on the Inter-X dataset. These results align with the premise of our work: in high-contact assistive tasks, although the recipient is incapable of performing the task independently, it is crucial for the recipient to actively learn how to coordinate and receive support, rather than merely replaying a motion. This bidirectional adaptation enables AssistMimic to handle such complex, force-exchanging interactions within physically natural motions. 

\item \textbf{Importance of Motion Prior Initialization}: Bootstrapping from the single-agent motion prior is essential for convergence. Without it, the RL process fails completely (0\% success rate) or succumbs to reward hacking—exploiting the objective function to produce unnatural, non-functional motions. (Note: Entries marked with $^{\dag}$ in Table~\ref{tab:expert-comparison} denote these reward-hacking failures). 
\item \textbf{Effectiveness of Reference Retargeting}: Dynamic reference retargeting is critical for maintaining valid physical contact, especially for the HHI-Assist dataset. Without it, the supporter blindly tracks the fixed reference trajectory, ignoring the recipient's dynamic deviations. This results in severe spatial misalignment, where the supporter fails to adjust its hands to the recipient's actual location, leading to missed contacts and failed assistance. This can also be observed in the zoomed-in regions highlighted by the red boxes in Figure~\ref{fig:hhi-assist-qualitative}. In the first row without retargeting, the supporter’s hands simply follow the reference trajectory and fail to adjust to the recipient’s actual position, resulting in inadequate support. Conversely, the module did not provide benefits on the Inter-X dataset. We hypothesize that this is because the recipients in Inter-X move dynamically over a much wider range than in HHI-Assist, which makes the relative goal position highly unstable.
\item \textbf{Robustness via Contact Promotion}: The contact-promotion reward drives the learning of active, force-aware interaction. By prioritizing physical contact over strict kinematic tracking, this term significantly enhances robustness to unseen recipient dynamics. It enables the policy to adapt to unexpected physical changes—such as increased mass, reduced PD gains ($k_p, k_d$), or torque limits—where a purely tracking-based policy would fail to exert sufficient supportive force.

\item \textbf{Enhanced Stability in Bed-based Assistance}: We observe that our method yields more stable assistance, as measured by a COM stability metric (Appendix~C). 
\end{itemize}

\subsection{Generalist Policy Evaluation} 
Table~\ref{tab:generalist-comparison} summarizes the tracking accuracy for generalist policies evaluated on the Inter-X dataset. When trained directly on a diverse subset of 30 interaction clips, \model{} achieves a success rate of 77.3\%, making it, to our knowledge, the first physics-based method to handle such a broad range of tightly coupled interactions.
Furthermore, we observe that distilling specialist policies into a single generalist agent via DAgger yields substantial performance gains. 
The top two rows of Figure~\ref{fig:teaser} present qualitative results on the Inter-X dataset obtained with a single generalist policy trained with \model{}. For additional qualitative results, please see Appendix~B.
\vspace{-2mm}

\subsection{Robustness to Unseen Recipient Dynamics}
\label{sec:constraints}
During training, we impose the physical constraints listed in Table~\ref{tab:phisical_limitations} on the recipient to enable learning of the caregiver’s support motions. 
To assess robustness to unseen recipient constraints, we evaluate zero-shot performance at test time.
As shown in Table~\ref{tab:expert-comparison}, \model{} exhibits strong robustness to unseen recipient constraints. Removing the contact reward or dynamic reference retargeting substantially degrades this robustness, suggesting that encouraging deliberate contacts during RL enables the policy to acquire diverse support strategies. Consequently, it handles changes in recipient limitations, such as unexpected stumbles or altered load distributions, with high accuracy.

\vspace{-2mm}

\subsection{Tracking of Generated Interactions}
\begin{figure}[t]
\centering
\centerline{\includegraphics[width=1.0\linewidth]{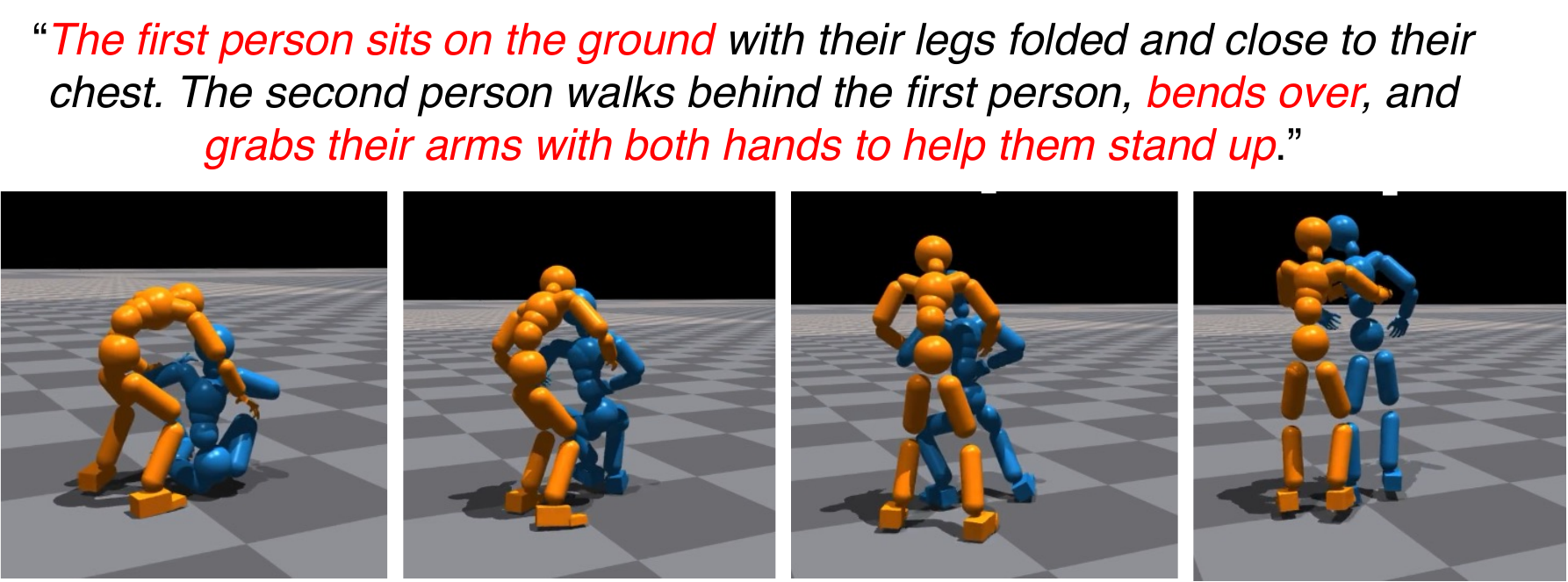}}
\vspace{-2mm}
\caption{Tracking results of \model{} on interaction trajectories generated by a motion diffusion model.}
\label{fig:generated_qualitative}
\vspace{-6mm}
\end{figure}

Figure~\ref{fig:generated_qualitative} illustrates the results of applying \model{} to trajectories synthesized by a text-conditioned motion diffusion model \cite{tevet2023human} (details in Appendix~F). 
The diffusion model was trained on "Help-up" interactions from the Inter-X dataset. During inference, we prompt it with unseen text descriptions from the same category. While the generated kinematics capture the interaction semantics, they often contain physical artifacts such as foot sliding and penetration.
{\model{}} successfully converts dense interaction kinematics into physically plausible motions. This result demonstrates that {\model{}} is effective not only for real interaction data but also for generated data, highlighting its generality and broad applicability.

\vspace{-2mm}

\begin{figure}[t]
\centering
\begin{minipage}[t]{0.35\linewidth}
    \centering
    \includegraphics[width=\linewidth]{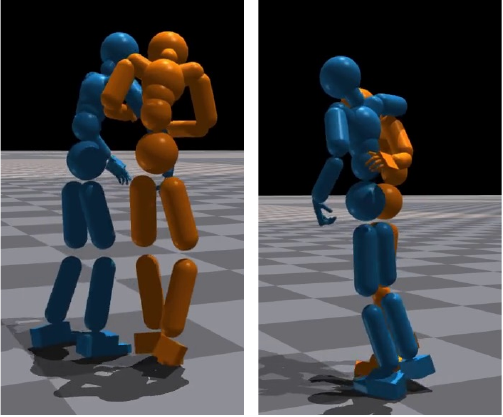}
    \vspace{-4mm}
    \subcaption{Generalization to unseen interactions.}
\end{minipage}
\hfill
\begin{minipage}[t]{0.55\linewidth}
    \centering
    \includegraphics[width=\linewidth]{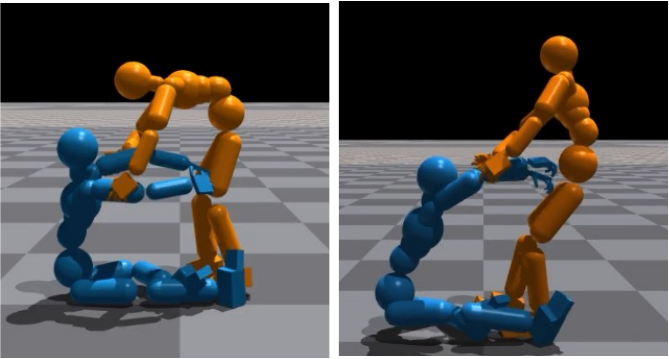}
    \vspace{-4mm}
    \subcaption{Failure cases due to limited hand dexterity.}
\end{minipage}
\vspace{-4mm}
\caption{Qualitative results: unseen interactions and failures.}
\label{fig:generalization_failure}
\vspace{-6mm}
\end{figure}

\subsection{Generalization to Unseen Supportive Actions}

As shown in Figure~\ref{fig:generalization_failure} (a), \model{} also generalizes to unseen interaction categories from Inter-X, such as \textit{support-by-arm}. 
Even though this interaction pattern is not observed during training, the humanoid successfully follows the motion while placing a supportive hand on the partner's arm and avoiding body collisions.

\subsection{Failure cases and limitations.} 
Figure~\ref{fig:generalization_failure} (b) shows failures in motions that require precise grasping and lifting of the recipient’s arms. 
These failures are primarily caused by limited hand dexterity, as grasping a human arm requires fine, coordinated finger control that is difficult to learn from noisy demonstrations and are further exacerbated by the recipient’s weight.
Tighter integration between motion planning and tracking is needed for more adaptive coordination. 
Incorporating visual observations is necessary to improve robustness to dynamic partner states. 
Transferring the learned behaviors from simulation to real humanoid systems remains an important challenge.

%% file: sec-cameraready/05_conclusion.tex
\section{Conclusion} 
We introduced \model{}, a multi-agent reinforcement learning framework for learning physics-aware, tracking controllers that reproduce and adapt to human–human interactions, including assistive behaviors. Unlike prior approaches that treat interaction partners independently or rely on fixed replayed trajectories, \model{} jointly learns policies for both agents, enabling reactive and physically consistent coordination. \model{} improves exploration and stability in RL through three key innovations:
(1) weight initialization from single-human tracking controllers, which stabilizes early learning;
(2) dynamic reference retargeting, allowing the policy to adjust to the recipient’s current pose; and
(3) contact-promoting rewards, encouraging physically meaningful interactions.
Experiments on the Inter-X and HHI-Assist show that \model{} is the first method capable of successfully tracking assistive interaction sequences. More broadly, our results demonstrate a generalizable framework for extending single-agent imitation learning to physically grounded multi-agent domains, with promising implications for assistive robotics, collaborative manipulation, and embodied simulation of social interactions.

%% file: supp-arxiv2.tex
\clearpage
\maketitlesupplementary
\section{Overview of the Supplementary Materials}
\label{sec:overview}
This supplementary document contains additional details and discussions of our \emph{AssistMimic}. Please also refer to the supplementary video \textcolor{magenta}{\texttt{assistmimic-video.mov}}, which provides an overview of our task setup and problem background, as well as motion-tracking visualizations of the learned assistive behaviors.
We highlight reference numbers associated with the main paper in \textcolor{blue}{blue}, and those associated with this supplementary document in \textcolor{red}{red}.

\section{Additional Qualitative comparison}
\label{sec:supp-qualitative}
\begin{figure*}[t]
\centering
\centerline{\includegraphics[width=1.0\linewidth]{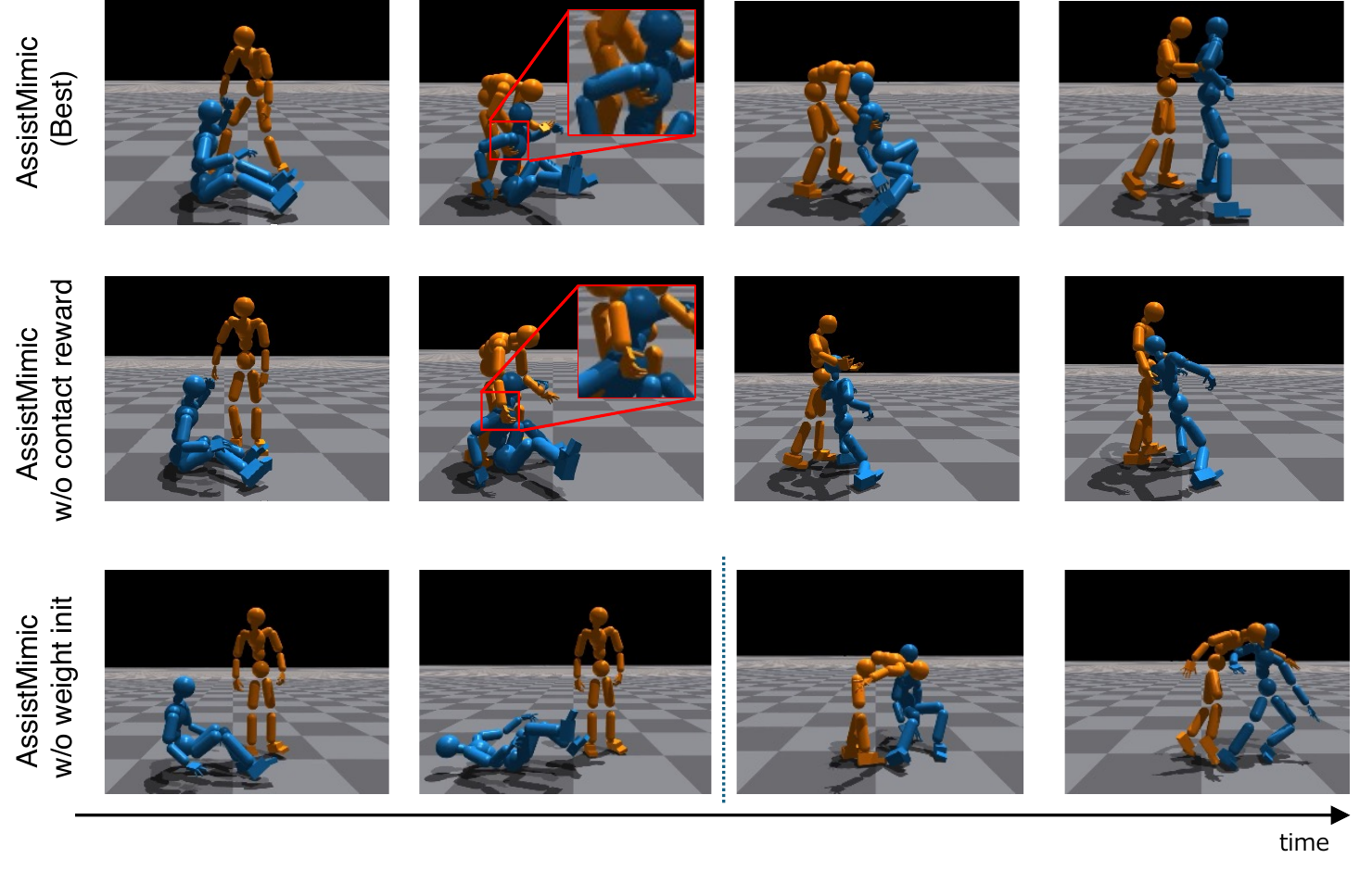}}
\vspace{-2mm}
\caption{The qualitative comparison of the specialist with Inter-X dataset. Supporter (orange) and Recipient (blue). Left to Right in chronological order.}
\label{fig:inter-x-qualitative-supp}
\end{figure*}

Figure~\ref{fig:inter-x-qualitative-supp} visualizes the ablation results of the specialist policies on the Inter-X dataset.  
Each row corresponds to a different variant, and frames are shown from left to right in chronological order.  
The first and second rows compare the best AssistMimic policy (w/o dynamic reference retargeting) and the variant without the contact-promoting reward on the \emph{same} help-up motion.  
The third row shows two separate examples for the variant without weight initialization, because the humanoids quickly lose balance and the episodes terminate early.

From the comparison between the first and second rows, especially in the zoomed-in regions highlighted by the red boxes, we can observe that introducing the contact-promoting reward enables the supporter to discover a correct assistive strategy even under noisy reference motions.  
In the second row, the supporter over-follows the noisy hand trajectory and ends up pressing down on the recipient from above, which leads to the recipient’s fall.  
In contrast, the best model maintains stable, supportive contact around the upper body and produces a more realistic help-up behavior.

For the variant without weight initialization (third row), training fails to progress: near-floor motions cause the characters to lose balance almost immediately in some cases (first and second columns), or, in other cases (third and fourth columns), the supporter approaches and reaches out but then leans its body weight onto the recipient, causing both agents to fall.

\begin{figure*}[t]
\centering
\centerline{\includegraphics[width=1.0\linewidth]{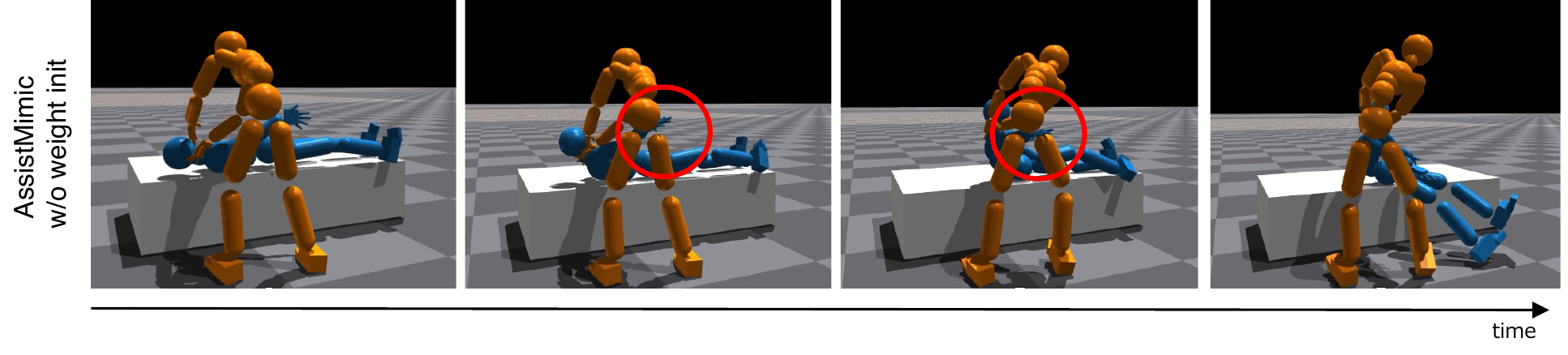}}
\vspace{-2mm}
\caption{The qualitative comparison with HHI-Assist dataset. Supporter (orange) and Recipient (blue).}
\label{fig:hhi-assist-supp}
\vspace{-4mm}
\end{figure*}

Figure~\ref{fig:hhi-assist-supp} shows the results on the HHI-Assist dataset when PHC is \emph{not} used for weight initialization.  
Compared to the behavior in \textcolor{blue}{Figure~4} of the main paper, the recipient here largely ignores the intended hand-tracking objective and instead touches the supporter’s waist, using the resulting reaction to lift its upper body.  
As also reflected by the low success rate in \textcolor{blue}{Table~3}, even the episodes that are counted as successful are dominated by such reward-hacking behaviors, indicating that meaningful assistive strategies are not learned in this setting.

\begin{table}[h]
\centering
\small
\vspace{-1mm}
\caption{Recipient COM standard deviation ($\downarrow$)} 
\vspace{-3mm}
\label{tab:recipient_com}
\setlength{\tabcolsep}{8pt}
\begin{tabular}{lccc}
\toprule
Model & Seen & Mass $\times 1.5$ & Max hip $\tau$ $\times 0.5$ \\
\midrule
\rowcolor{pink!30}
Ours              & \textbf{0.0921} & \textbf{0.0738} & 0.0865 \\
(-) Dyn Retar           & 0.1038 & 0.0902 & 0.0924 \\
~~(-) Cont   & 0.0938 & 0.0838 & \textbf{0.0849} \\
\bottomrule
\end{tabular}
\vspace{-2mm}
\end{table}

\section{COM Stability Metric}
\label{sec:detail-com}

To quantify the stability of assistive interactions, we measure the temporal variation of the recipient's center of mass (COM) during each episode.

\paragraph{Definition.}
Let $\mathbf{c}_t \in \mathbb{R}^3$ denote the 3D position of the recipient's COM at timestep $t$, computed from the mass-weighted average of all body segments. 
We define the COM stability metric as the standard deviation of $\mathbf{c}_t$ over time:
\begin{equation}
\sigma_{\text{COM}} = \sqrt{\frac{1}{T} \sum_{t=1}^{T} \left\| \mathbf{c}_t - \bar{\mathbf{c}} \right\|^2}, \quad 
\bar{\mathbf{c}} = \frac{1}{T} \sum_{t=1}^{T} \mathbf{c}_t,
\end{equation}
where $T$ is the number of timesteps in the episode.

\paragraph{Evaluation protocol.}
To ensure a fair comparison, we compute $\sigma_{\text{COM}}$ only on motion sequences where all compared methods successfully complete the task. 
This avoids bias due to early termination or failure cases, which would otherwise result in shorter trajectories and artificially lower variance.

\paragraph{Interpretation.}
A lower $\sigma_{\text{COM}}$ indicates more stable assistance, as the recipient's body is maintained with less oscillation or unintended movement. 
This metric is particularly relevant for bed-based caregiving scenarios (e.g., HHI-Assist), where maintaining a steady posture is critical.

\paragraph{Results.}
As summarized in Table~\ref{tab:recipient_com}, AssistMimic achieves lower recipient COM standard deviation compared to the ablated variants on the HHI-Assist when evaluated on motions successfully completed by all policies. Together with the results in \textcolor{blue}{Table~3}, this indicates that our approach achieves both high success rates and stable assistance.

\section{Additional implementation details}
\label{sec:additional_implementation_details}
\paragraph{Optimization setup.}
For the multi-agent RL optimization, we use Proximal Policy Optimization (PPO)~\cite{schulman2017proximal} for both the supporter and the recipient policies.  
For the specialist comparisons in \textcolor{blue}{Table~2, 3}, policies are trained for 2k iterations, which we found sufficient for these focused subsets.
For the generalist evaluation in \textcolor{blue}{Table~4}, our AssistMimic w/o DAgger (1st line) is trained for $9 \text{k}$ iterations. Similarly, we train specialist teacher policies for $9 \text{k}$ iterations to provide high-fidelity supervision, from which the final generalist policy is distilled via DAgger~\cite{ross2011reduction} over $4 \text{k}$ epochs.
The learning rate for both agents is set to $5\times 10^{-6}$, and we apply a step decay by a factor of $0.1$ once training reaches $600$ PPO iterations.  
The critic network is implemented as an MLP that takes the same input as the actor, augmented with a role label indicating whether the agent is the \emph{recipient} or the \emph{caregiver} (supporter). 

\paragraph{Motion prior and fine-tuning.}
AssistMimic inherits weights from a single-human PHC tracking controller~\cite{Luo_2023_ICCV} through our weight-initialization scheme.  
However, motions of the recipient near the floor are largely out-of-domain for the original PHC model trained on AMASS~\cite{mahmood2019amass}.  
To address this, we perform a small-scale fine-tuning stage using only recipient motion data.  
After this stage, we filter out clips that can already be successfully tracked by the single-human PHC rollout alone, since such motions do not require meaningful support and should not be treated as assistive examples. Unlike the original PHC work, which trains multiple motion primitives, all of our experiments are conducted with a \emph{single} primitive.

\paragraph{Generalist training.}
As discussed in the ablation study in the main paper, the dynamic reference retargeting module did not provide benefits on the Inter-X dataset.  
Therefore, for training the Inter-X generalist policy, we disable this module and first train four specialist policies, each on a subset of Inter-X help-up motions.  
We then perform online policy distillation using DAgger~\cite{ross2011reduction}, where the generalist policy is trained to imitate the actions of these specialists.  
For this distillation stage, we optimize the generalist for 1k epochs using a squared-error loss between the generalist and specialist actions.

\paragraph{Evaluation protocol for the HHI-Assist bed setting.}
\label{sec:detail-hhi-eval}
In many HHI-Assist bed sequences, the supporter steps away from the recipient near the end of the motion after the assistance is completed. 
Because the recipient is no longer supported during this phase, falls occurring in this stage should not be considered assistive failures. 
Thus, for quantitative evaluations (\textcolor{blue}{Table 3}, Table~\ref{tab:recipient_com}), we discard the final 15 frames of each 30\,FPS motion clip and compute all metrics on the remaining frames.

\paragraph{Caregiving on a chair.}
The chair-based caregiving behaviors shown in \textcolor{blue}{Figure~1} of the main paper are trained using reference motions from the HHI-Assist dataset.  
During this training, we apply the same physical limitations to the recipient as those listed in \textcolor{blue}{Table~1}, analogous to the bed-care setting.

\section{Detailed reward breakdown}
In this section, we describe the reward details that could not be included in the main paper due to space limitations.

\subsection{Overall reward structure}
\label{sec:reward-details}
  Our method builds upon the Adversarial Motion Priors (AMP)~\cite{peng2021amp} framework.
  The per-timestep reward for agent $m\in\{S,R\}$ (supporter/recipient) combines
  a task reward and a discriminator reward:
  \begin{equation}
    r^{(m)}_{t} = \lambda_{\text{task}}\, r^{(m)}_{\text{task},t}
                + \lambda_{\text{disc}}\, r_{\text{disc},t},
    \label{eq:amp_reward}
  \end{equation}
  where $r_{\text{disc},t}$ is the AMP discriminator reward that encourages
  physically plausible motion, and $\lambda_{\text{task}}, \lambda_{\text{disc}} > 0$
  are scalar weights (set to $0.5$ each in our experiments).

  \subsection{Task reward decomposition}
  The task reward is further decomposed into tracking, power penalty, and
  assistive terms:
  \begin{equation}
    r^{(m)}_{\text{task},t}
    = \lambda_{\text{track}}\, r^{(m)}_{\text{track},t}
    + \lambda_{\text{power}}\, r^{(m)}_{\text{power},t}
    + \lambda_{\text{assist}}\, r^{(m)}_{\text{assist},t},
    \label{eq:task_reward}
  \end{equation}
  where $\lambda_{\text{track}},\lambda_{\text{power}},\lambda_{\text{assist}}>0$
  are scalar weights.

  \textbf{Tracking reward.}
  The tracking reward measures how closely agent $m$ follows its reference motion:
  \begin{equation}
    r^{(m)}_{\text{track},t}
    = \frac{1}{J}\sum_{j=1}^{J} \exp\!\bigl(-k_{\text{track}} \cdot d_j(\hat{\mathbf{q}}^{(m)}_{j,t}, \mathbf{q}^{(m)}_{j,t})\bigr),
    \label{eq:tracking_reward_detailed}
  \end{equation}
  where $J$ is the number of joints, $d_j(\cdot,\cdot)$ is the distance metric
  for joint $j$ (combining position and rotation differences),
  $\hat{\mathbf{q}}^{(m)}_{j,t}$ is the current state,
  $\mathbf{q}^{(m)}_{j,t}$ is the reference state, and $k_{\text{track}}$
  is a scaling factor.

  \textbf{Power penalty.}
  The power penalty discourages excessive joint actuation by penalizing
  the mechanical power:
  \begin{equation}
    r^{(m)}_{\text{power},t} = -\lambda^{(m)}_{\text{power}} \sum_{j=1}^{J} |\tau^{(m)}_{j,t} \cdot \dot{q}^{(m)}_{j,t}|,
    \label{eq:power_penalty}
  \end{equation}
  where $\tau^{(m)}_{j,t}$ is the torque applied to joint $j$ at time $t$,
  $\dot{q}^{(m)}_{j,t}$ is the joint velocity, and
  $\lambda^{(S)}_{\text{power}}=0.0015$ for the supporter and
  $\lambda^{(R)}_{\text{power}}=0.002$ for the recipient.

  \textbf{Assistive reward.}
  The assistive term encourages effective support:
  \begin{equation}
    r^{(m)}_{\text{assist},t}
    = \alpha_{\text{head}}\, r_{\text{head},t}^{(R)}
    + \alpha_{\text{torque}}\, r_{\text{torque},t}^{(R)},
    \label{eq:assist_reward}
  \end{equation}
  where $r_{\text{head},t}^{(R)} = \min(h^{(R)}_t, h^{(R)}_{\max})$
  rewards a higher head height of the recipient (clamped by the maximum
  achievable height $h^{(R)}_{\max}$), and
  $r_{\text{torque},t}^{(R)} = \exp(-\|\boldsymbol{\tau}^{(R)}_t\|_1 / \sigma_{\text{torque}})$
  rewards reductions in the recipient's joint torques.

  \subsection{Caregiver -- recipient coupling}
  We first compute the per-agent rewards $\tilde r^{(S)}_{t}$ and
  $\tilde r^{(R)}_{t}$ according to Eqs.~\eqref{eq:amp_reward}--\eqref{eq:assist_reward}.
  The final rewards used for optimization are then given by
  \begin{equation}
    r^{(R)}_{t} = \tilde r^{(R)}_{t}, \qquad
    r^{(S)}_{t} = \tfrac{1}{2}\,\tilde r^{(S)}_{t}
                + \tfrac{1}{2}\,\tilde r^{(R)}_{t},
    \label{eq:caregiver_recipient_coupling}
  \end{equation}
  so that the caregiver (supporter) is explicitly encouraged to maximize
  not only its own reward but also the overall reward of the recipient.

Table~\ref{tab:reward-retargeting-hparams} summarizes the hyperparameters for the proposed contact-promoting reward and dynamic reference retargeting, as well as the additional reward terms introduced in this section.

\begin{table*}[t]
\centering
\small
\setlength{\tabcolsep}{4pt}
\begin{tabular}{llp{8.0cm}l}
\toprule
Category & Symbol & Description & Value \\
\midrule
Overall reward &
$\lambda_{\text{task}}$ &
Weight for task reward $r^{(m)}_{\text{task},t}$ in the per-timestep reward & 0.5
\\
&
$\lambda_{\text{disc}}$ &
Weight for discriminator reward $r_{\text{disc},t}$ from AMP. & 0.5
\\
\midrule
Task reward decomposition &
$\lambda_{\text{track}}$ &
Weight for tracking term 
& 1.0 \\
& 
$\lambda_{\text{assist}}$ &
Weight for assistive term $r^{(m)}_{\text{assist},t}$. & 1.0
\\
&
$k_{\text{track}}$ &
Scaling factor in the exponential tracking reward 
& 100
\\
&
$\lambda^{(S)}_{\text{power}}$ &
Power penalty coefficient for the supporter 
& 0.0015 
\\
&
$\lambda^{(R)}_{\text{power}}$ &
Power penalty coefficient for the recipient 
& 0.002
\\
\midrule
Assistive reward &
$\alpha_{\text{head}}$ &
Weight for head-height term $r^{(R)}_{\text{head},t}$ in the assistive reward $r^{(m)}_{\text{assist},t} = \alpha_{\text{head}} r^{(R)}_{\text{head},t} + \alpha_{\text{torque}} r^{(R)}_{\text{torque},t}$. & 1.0
\\
&
$\alpha_{\text{torque}}$ &
Weight for torque-reduction term $r^{(R)}_{\text{torque},t}$. &
0.0 (Inter-X)
\\
& & & 0.5 (HHI-Assist)
\\
&
$h^{(R)}_{\max}$ &
Normalization constant for recipient head height in $r^{(R)}_{\text{head},t} = \min(h^{(R)}_t / h^{(R)}_{\max}, 1.0)$. & 2.0
\\
&
$\sigma_{\text{torque}}$ &
Scaling factor in the torque reduction term $r^{(R)}_{\text{torque},t} = \exp(-\|\tau^{(R)}_t\|_1 / \sigma_{\text{torque}})$. & 150
\\
\midrule
Caregiver--recipient coupling &
$\tfrac{1}{2}$ (mixing weight) &
Mixing coefficient in the final supporter reward $r^{(S)}_t = \tfrac{1}{2} \tilde{r}^{(S)}_t + \tfrac{1}{2} \tilde{r}^{(R)}_t$. &
\\
\midrule
Dynamic hand reference retargeting &
$\tau_{\text{dist}}$ &
Distance threshold for activating reference retargeting in the gating term $G_t = \mathbb{I}(\|p^{(S)}_{\text{root},t} - p^{(R)}_{\text{root},t}\|_2 \le \tau_{\text{dist}})$. & 1.3
\\
\midrule
Contact-promoting reward (supporter) &
$d_{\text{th}}$ &
Distance threshold for proximity indicator $\chi_{i,t} = \mathbb{I}(d_{i,t} \le d_{\text{th}})$, where $d_{i,t}$ is the minimum distance between supporter wrist $i$ and recipient upper-body joints. & 0.4
\\
&
$\alpha$ &
Distance scaling factor in the contact term $\beta f_{i,t} \exp(-\alpha d_{i,t})$. & 2.5
\\
&
$\beta$ &
Force scaling factor in the contact term $\beta f_{i,t} \exp(-\alpha d_{i,t})$. & 0.5
\\
&
$b_{\text{contact}}$ &
Sparse bonus added when contact is established in the contact-promoting reward. & 0.05
\\
&
$f_{\text{th}}$ &
Force threshold in the saturated contact-force aggregation $f_{i,t} = \sum_{\ell \in H_i \setminus \{i\}} \min(\exp(\|f_{\ell,t}\|_2 - f_{\text{th}}), 1)$. & 1.0
\\
\bottomrule
\end{tabular}
\caption{Hyperparameters used in the reward design and dynamic hand reference retargeting.}
\label{tab:reward-retargeting-hparams}
\end{table*}

\section{Detailed information about the diffusion planner}
\label{sec:diffusion}

\subsection{Overview}
Our diffusion planner is a denoising diffusion transformer that auto-regressively generates joint positions for both the supporter and the recipient, conditioned on text descriptions. During inference, we prompt the model with unseen descriptions from the "Help-up" category to synthesize novel motion sequences, which are subsequently tracked using AssistMimic.

\subsection{Data representation}
We train the planner using motion sequences from the Inter-X dataset labeled as "Help-up" (a small subset is held out for testing). Each sequence contains SMPL-X parameters for both individuals along with a corresponding text description. These sequences are converted into a representation suitable for diffusion-based generation.

Before constructing the motion representation, each motion sequence is preprocessed to ensure a canonical orientation. Specifically, the average displacement vector between the root joints of the supporter and recipient over the sequence is computed, and the entire motion is rotated such that this vector is aligned with the $x$-axis.

For each individual at frame $i$ (subscripts $c$ for supporter and $r$ for recipient), we extract the joint positions in the global-frame via forward kinematics:
\[
\mathbf{p}_{c}^{i},\, \mathbf{p}_{r}^{i} \in \mathbb{R}^{22 \times 3}.
\]

Joint velocities are computed using finite differences:
\[
\mathbf{v}_{c}^{i} = \mathbf{p}_{c}^{i} - \mathbf{p}_{c}^{i-1}, \qquad
\mathbf{v}_{r}^{i} = \mathbf{p}_{r}^{i} - \mathbf{p}_{r}^{i-1} \in \mathbb{R}^{22 \times 3}.
\]

The left and right-hand SMPL-X axis-angle poses are converted to the continuous 6D representation, yielding
\[
\mathbf{h}_{c,\text{L}}^{i},\, \mathbf{h}_{c,\text{R}}^{i},\,
\mathbf{h}_{r,\text{L}}^{i},\, \mathbf{h}_{r,\text{R}}^{i}
\in \mathbb{R}^{15 \times 6}.
\]

Thus, the motion feature vector at frame $i$ for each individual is
\[
\mathbf{x}_c^{i} =
\big[
\mathbf{p}_c^{i},\;
\mathbf{v}_c^{i},\;
\mathbf{h}_{c,\text{L}}^{i},\;
\mathbf{h}_{c,\text{R}}^{i}
\big]
\in \mathbb{R}^{312},
\]
\[
\mathbf{x}_r^{i} =
\big[
\mathbf{p}_r^{i},\;
\mathbf{v}_r^{i},\;
\mathbf{h}_{r,\text{L}}^{i},\;
\mathbf{h}_{r,\text{R}}^{i}
\big]
\in \mathbb{R}^{312}.
\]

Concatenating both individuals produces the full representation:
\[
\mathbf{x}^{i} =
\big[
\mathbf{x}_c^{i},\;
\mathbf{x}_r^{i}
\big]
\in \mathbb{R}^{624}.
\]

A motion window is defined as
\[
\mathbf{x}^{i:i+N}
=
\{ \mathbf{x}^{i}, \mathbf{x}^{i+1}, \dots, \mathbf{x}^{i+N-1} \}.
\]

The accompanying text description is encoded using a frozen DistilBERT encoder~\cite{Sanh2019DistilBERTAD}, producing
\[
\mathbf{z}_{\text{text}} \in \mathbb{R}^{N_{\text{tokens}} \times 768}.
\]

\subsection{Model architecture}
Our implementation of the diffusion planner $G$ is based on MDM~\cite{tevet2023human}. It is trained to predict the clean motion window $\hat{\mathbf{x}}_{0}^{i:i+N}$ from a noisy version $\mathbf{x}_{t}^{i:i+N}$, conditioned on the previous window and the text embedding. The diffusion timestep is denoted by $t$:
\[
\hat{\mathbf{x}}_{0}^{i:i+N}
=
G\big(
\mathbf{x}_{t}^{i:i+N},
\, t,
\, \mathbf{x}^{i-N:i},
\, \mathbf{z}_{\text{text}}
\big).
\]

During training, the windows $\mathbf{x}^{i-N:i}$ and $\mathbf{x}_{t}^{i:i+N}$ are concatenated and processed with self-attention, while the text embedding is incorporated via cross-attention in the transformer decoder layers. The motion windows along with the text embedding are all projected to the same latent dimension to be compatible with the attention layers.

We optimize the standard $\mathcal{L}_{\text{simple}}$ loss together with a loss for temporal consistency $\mathcal{L}_{\text{vel}}$, weighted by $\lambda_{\text{vel}} = 25$:
\[
\mathcal{L}_{\text{vel}}
=
\frac{1}{N-1}
\sum_{i=1}^{N-1}
\left\|
    \left( \mathbf{x}_{0}^{i+1} - \mathbf{x}_{0}^{i} \right)
    -
    \left( \hat{\mathbf{x}}_{0}^{\,i+1} - \hat{\mathbf{x}}_{0}^{\,i} \right)
\right\|_2^{2}.
\]

\[
\mathcal{L}
=
\mathcal{L}_{\text{simple}}
+
\lambda_{\text{vel}}\, \mathcal{L}_{\text{vel}}.
\]

Specifically, our model uses a transformer decoder with 10 layers and 8 attention heads, a feedforward dimension of 2048, and a latent dimension of 768, with dropout set to 0.1. The prediction window is $N = 20$ frames. During training, the conditioning variables are masked with a probability of 0.1 to enable unconditional generation. The diffusion process uses 100 timesteps with a cosine beta schedule. We optimize the model using AdamW with a learning rate of $2\times10^{-4}$, betas $(0.9, 0.95)$, and weight decay $10^{-3}$. Training is performed for 8 hours on an NVIDIA A6000 GPU.

\subsection{Sampling}
During inference, the model is given a previously unseen text description of the intended motion and auto-regressively generates 180 frames in chunks of 20 frames. Each subsequent window is conditioned on the text description and the 20 frames generated in the previous step. The first 20 frames are generated using only the provided text as conditioning.

Finally, the SMPL-X parameters are recovered from the diffusion output via inverse kinematics using VPoser~\cite{SMPL-X:2019}. This generated motion sequence is then tracked using AssistMimic.